\documentclass[12pt]{article}
\usepackage{arxiv}

\usepackage[utf8]{inputenc} 
\usepackage[T1]{fontenc}    
\usepackage{hyperref}       
\usepackage{url}            
\usepackage{booktabs}       
\usepackage{amsfonts}       
\usepackage{nicefrac}       
\usepackage{microtype}      
\usepackage{graphicx}
\usepackage{amsmath,amssymb}
\usepackage{enumitem}
\usepackage{subcaption}
\usepackage{comment}
\usepackage{epsfig}
\usepackage{float}
\usepackage{color}
\newcommand{\ra}[1]{\renewcommand{\arraystretch}{#1}}

\title{Perceiving Traffic from Aerial Images}

\author{
  George Adaimi\\
  VITA, EPFL\\
  Switzerland\\
  \texttt{george.adaimi@epfl.ch} \\
   \And
 Sven Kreiss \\
  VITA, EPFL\\
  Switzerland\\
  \texttt{sven.kreiss@epfl.ch} \\
  \And
  Alexandre Alahi\\
  VITA, EPFL\\
  Switzerland\\
  \texttt{alexandre.alahi@epfl.ch} \\
}
\begin{document}

\maketitle

\begin{abstract}
Drones or UAVs, equipped with different sensors, have been deployed in many places especially for urban traffic monitoring or last-mile delivery. It provides the ability to control the different aspects of traffic given real-time obeservations, an important pillar for the future of transportation and smart cities. With the increasing use of such machines, many previous state-of-the-art object detectors, who have achieved high performance on front facing cameras, are being used on UAV datasets. When applied to high-resolution aerial images captured from such datasets, they fail to generalize to the wide range of objects' scales. In order to address this limitation, we propose an object detection method called Butterfly Detector that is tailored to detect objects in aerial images. We extend the concept of fields and introduce butterfly fields, a type of composite field that describes the spatial information of output features as well as the scale of the detected object. To overcome occlusion and viewing angle variations that can hinder the localization process, we employ a voting mechanism between related butterfly vectors pointing to the object center. We evaluate our Butterfly Detector on two publicly available UAV datasets (UAVDT and VisDrone2019) and show that it outperforms previous state-of-the-art methods while remaining real-time.
\end{abstract}


\section{Introduction}

Smart cities have been increasing around the world especially with the increasing importance of sustainable development and the effect of AI on our everyday life. It is projected also that the market for such cities will keep on growing~\cite{smartCities_2020}. For a successful implementation of smart cities, a large amount of data is required to better understand its different aspects such as congestion and other transportation network conditions. Visual information have long been collected through the use of many static cameras placed at different locations. Installing such cameras with the required infrastructure is costly and fails to monitor a very large area. Moreover, certain unexpected scenarios require visual information at un-monitored location, such as in the case of huge congestions caused by an accident. As a result, data collected from unmanned aerial vehicles (UAV) or drones have been increasing in popularity for urban traffic monitoring. One such example is the introduction of a new drone-collected large-scale dataset called pNEUMA
~\cite{barmpounakis2020new} which aims to help researchers better study and model traffic congestion. Collecting such large datasets requires analyzing large amounts of objects across many images. This work was traditionally outsourced to manual labellers such as Amazon Mechanical Turk (MTurk)
~\cite{amazon_turk}. With the huge success of deep learning, especially in the field of computer vision, most of the labelling work has become automated. In addition to using UAVs for dataset collection, many people believe that drones is the solution to increase the efficiency of last-mile deliveries~\cite{wang2019vehicle,cheng2020drone,gonzalez2020truck,SALAMA2020620,MURRAY2020368}. This requires the drone to be able to visually understand its surrounding using different sensors. Thus, paradigms for different vision tasks are still challenged with an arms race of methods every year. One of the most prominent vision tasks is \textit{object detection}. From traditional methods such as sliding window and deformable parts model (DPM)~\cite{felzenszwalb2009object} to deep neural networks with different components
~\cite{Fan_2020_CVPR}, the object detection race is still on.

\begin{figure}[t]
\begin{center}
\includegraphics[width=0.55\linewidth]{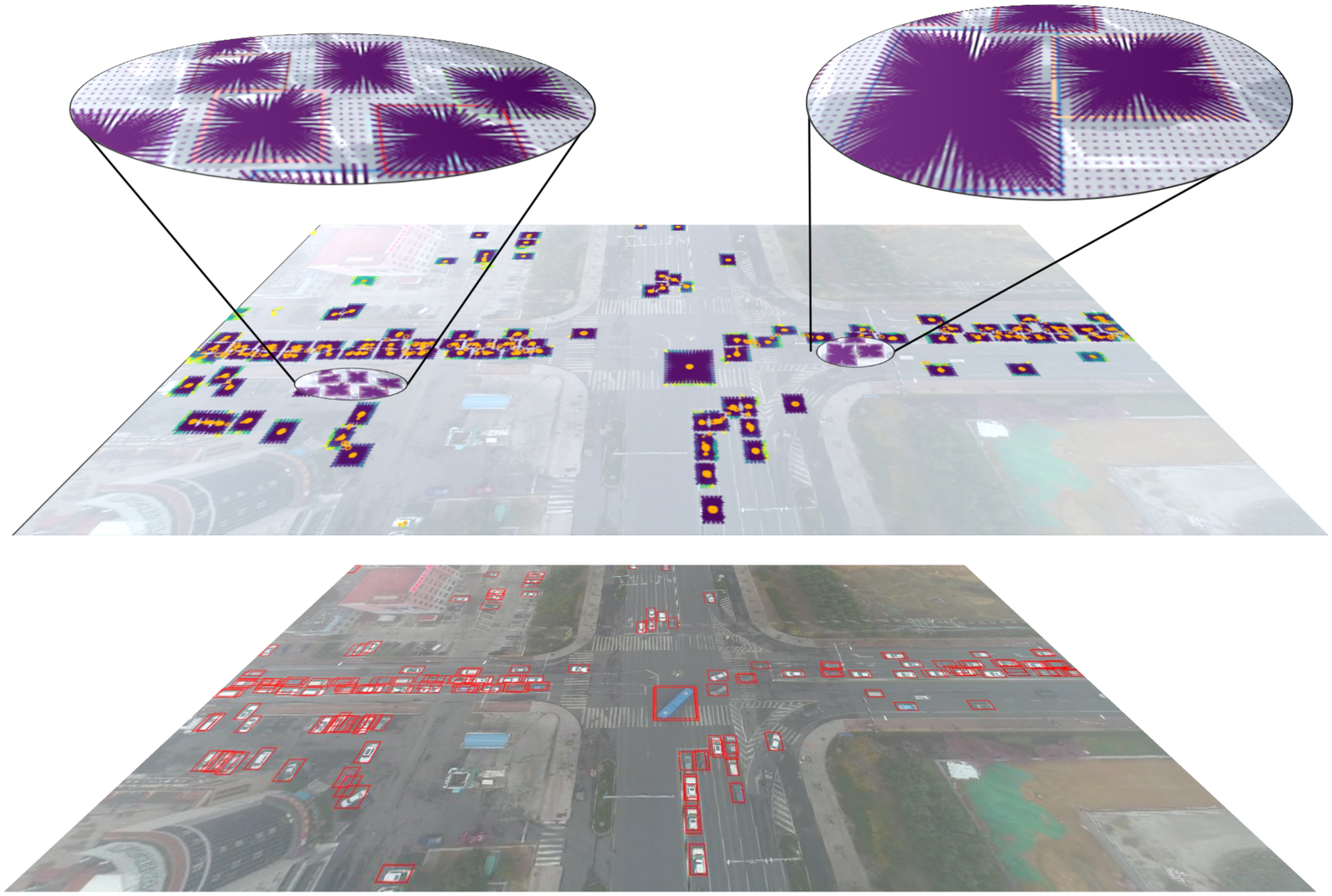}

\end{center}
\caption{
Butterfly fields outputted by our method overlayed on top of an aerial image (bottom). The purple vector components are pointing towards the center (orange circle). Using the different components of our butterfly fields allows us to better detect and localize the vehicles in the image}
\label{fig:pull}
\end{figure}

Although general object detectors have achieved promising results on front facing cameras \cite{law2018cornernet,zhou2019objects}, they fail to perform as well when used from a drone/UAV point-of-view. Aerial images capture a large scene made of hundreds of objects occupying a tiny portion of the image with very limited number of pixels (Figure~\ref{fig:pull}). Hence, this adds an exciting challenge to the problem of object detection. In this work, we propose a method tailored to detect objects in aerial images. Our method targets applications such as traffic analysis, smart cities, and more broadly digital twins to name a few.

With the prevalence of UAVs with high resolution cameras, two new datasets were recently released called UAVDT~\cite{du2018unmanned} and VisDrone2019~\cite{zhuvisdrone2018}. A common practice is to re-train and adapt some of the design choices of existing object detectors (\textit{e.g.}, Yolo \cite{redmon2016you}) to the new distribution of the data. For instance, Wu et al.~\cite{wu2019delving} adapted the sizes of the anchor boxes of Faster R-CNN \cite{ren2015faster} to improve its results on aerial images. Ding et al.~\cite{Ding_2019_CVPR} also introduced an ROI transformer to replace the conventional ROI and deal with a common challenge in UAV images, variations in objects' orientations. Yet, state-of-the-art methods still suffer from detecting tiny objects of any rotation.

Recent object detectors, using deep neural networks, can be categorized as anchor-based approaches and anchor-free approaches. On one hand, anchor-based methods leverage strong priors on scale and aspect ratio by pre-defining specific bounding boxes as initial detections. Rather than directly detecting the bounding boxes, the objects are located relative to pre-defined boxes obtained by analyzing the training dataset. Hence, they do not generalize well to objects of different scales or to images of different resolutions, especially when used with different datasets. On the other hand, anchor-free methods focus on locating corners and centers of objects directly with the use of prior knowledge. They are very challenging to train due to the sparsity of their output feature map. This is especially true in the case of objects in high-resolution images, a common characteristic of aerial images.

To overcome the limitations of both anchor-based and anchor-free approaches, we introduce a novel detector,  called Butterfly Detector (BD), that extends and adapts field theory, which has been gaining interest in the 2D human pose estimation task \cite{cao2018openpose,kreiss2019pifpaf,papandreou2018personlab}. In previous anchor-free models, only a specific feature in the output feature map of a model is responsible to localize the center or corner of an object. In our work, we make use of fields, which we call \textit{butterfly fields}, to localize the center and predict the width and height of an object. Each output feature that contains information about the object has a set of field components spatially pointing towards the center. Rather than implementing a naive approach of using all predicted bounding boxes at every output feature, we implement a voting mechanism leveraging the direction and information of related points in a field to improve localization as well as obtain a significant reduction in false positives. We also improved our precision, especially for small objects, by relying on a sub-pixel localization. Our experiments show that we outperform previous state-of-the-art methods on UAVDT and VisDrone2019 dataset. We especially achieve a boost in performance on images that are recorded at high altitudes (Table~\ref{table:sot_uavdt}) showcasing the importance of our method on aerial images. The code will be made available\footnote[1]{https://github.com/vita-epfl/butterflydetector}. Our method can also be tested using the video-labelling tool,
UltimateLabeling\footnote[2]{https://github.com/alexandre01/UltimateLabeling}, where it is used to detect vehicles.

\section{Related Work}

Object detection can be done at 2D scale~\cite{ren2015faster,He_2017_ICCV,redmon2016you,liu2016ssd} or 3D scale~\cite{Meyer_2019_CVPR,Qi_2019_ICCV,Li_2019_CVPR,Liu_2019_CVPR}. Since we are dealing with aerial images, we focus on 2D detection and further divide these methods into traditional detectors, anchor-based and anchor-free deep learning detectors.

\subsection{Traditional detectors for aerial images}
During the rise of satellites and aerial imagery, a lot of effort was put in developing methods to detect different objects~\cite{gupte2002detection}. This was quite challenging because of many different variation such as the small object size and different weather conditions. One of the earliest form of object detection relied on using a hand-crafted template per class and measuring its similarity at every location in the image using the sliding window technique. A lot of advancements have been been done to improve the templates used, from rigid templates
~\cite{chaudhuri2012semi,weber2012spatial} to deformable ones
~\cite{lin2014rotation,ahmadi2010automatic}. However, these methods are difficult to generalities ze to different shapes and orientations. To solve this, researchers moved toward extracting different features that help in localization and classification, such as color intensities and color gradients. One of the most interesting work in detection was done by ~\cite{dalal2005histograms}
 where they introduced histogram of oriented gradients (HoG) for human detection. The features extracted from the image are the distribution of oriented gradients, which allows the detection of edges. Many work since then extended this idea and further improved it to deal with different variations that occur while detecting objects. For instance, rotation-invariant HoG features~\cite{liu2014rotation} were developed to make these features resilient to rotational changes while deformable HoG~\cite{almazan2013deformable} were introduced to be resilient against occlusion and shape changes. Although pioneering works relied on handcrafted features and often dealt with satellite images \cite{cheng2013object,cheng2014multi,eikvil2009classification,grabner2008line,zhang2015generic,zhang2014nonlinear}, we will focus on reviewing recent deep learning based methods tailored for UAV images.

\subsection{Anchor-based detector for UAV images}

When AlexNet~\cite{krizhevsky2012imagenet} was first used for vehicle detection in aerial images, it was able to surpass all traditional methods. Since then, deep convolutional neural networks have dominated object detection with the release of Faster R-CNN~\cite{ren2015faster}, Mask R-CNN~\cite{He_2017_ICCV}, YOLO~\cite{redmon2016you}, and SSD~\cite{liu2016ssd}. Although these methods achieved good performance on natural images (MS COCO~\cite{lin2014microsoft} and PASCAL VOC~\cite{everingham2010pascal}), they perform poorly on high-resolution aerial images. Since most of these methods rely on anchor boxes and objects that are rather big, they require a lot of modifications to improve their performance on a different task such as UAV object detection.

Rather than re-inventing the wheel, many works have extended Faster R-CNN~\cite{ren2015faster}, R-FCN~\cite{dai2016r}, SSD~\cite{liu2016ssd}, and YOLO~\cite{redmon2016you} for vehicle detection~\cite{sommers2018,sommer2017fast,Li2019CCIGT,sommer2018multi,yang2018deep,ringwald2019uav} by proposing many adaptations to overcome challenges characteristic of aerial images. These adaptations include changing the number and size of anchor boxes, increasing the resolution, as well as making use of skip connections in the network. Tang et al.~\cite{tang2017vehicle} proposed replacing the RPN module of Faster R-CNN with a hyper region proposal network (HRPN) that makes use of hierarchical feature maps. Deng et al.~\cite{deng2017toward} extended this work by learning vehicle attributes as well as location and type.

Wu et al.~\cite{wu2019delving} proposed augmenting an off-the-shelf detector to also learn different attributes for every input such as altitude, weather, and viewing angle. This is done in an adverserial manner to disentangle these variations from the extracted features. Another work by Cai et al.~\cite{cai2019guided} uses an anchor-free architecture with background and foreground attention modules as well as multi-level features to improve object detection. In order to deal with small objects, Duan et al.~\cite{duan2019detecting} introduced a channel-aware de-convolution layer to exploit features from multiple channels of different layers. They also developed a Multi-RPN module that performs multiple detection at different layers simultaneously. In contrast, Li et al.~\cite{li2019simultaneously} combined bottom-up and top-down attention mechanisms to extract more discriminative features. By also counting the number of objects in a scene, they were able to show that these two tasks aid each other and lead to a boost in performance. Other methods utilized clustering~\cite{abs_1904_08008}, feature fusion network~\cite{long2019object}, or rotation-invariant cascaded forest~\cite{ma2019vehicle} to perform vehicle detection. All these methods, make use of either extra annotations, attention, or the fusion of multiple features from different layers to deal with different scales. This makes both the training and inference more complicated and slow.


\subsection{Anchor-free Object Detectors}
Due to the drawbacks of anchor-based object detectors, there has been recent work focusing on developing anchor-free detectors. Such methods detect the corners and centers of bounding boxes by outputting a heatmap over the image highlighting their location~\cite{tychsen2017denet,wang2017point,law2018cornernet,law2019cornernet,zhou2019bottom,zhou2019objects}. Point Linking Network (PLN)~\cite{wang2017point} detects the four corners and centers of bounding boxes. Then each corner predicts which output feature cell has a high probability of being the center. Then a bounding box is formed from each pair of center and corner. PLN is different then our Butterfly detector since we do not only make use of corners but rather every point inside the bounding box. Moreover, instead of predicting a probability over all feature cells, each point predicts a vector pointing to the center, allowing us to have a floating point precision of the center location. Law et al.~\cite{law2018cornernet,law2019cornernet}, on the contrary, predict only two opposite corners and use associative embedding to group them together. Since corners usually lie outside an object, they might not contain a lot of appearance features. In order to remedy this problem, Zhou et al.~\cite{zhou2019bottom} suggested using points that lie at the extreme edges of an object. They were able to show promising results, but this limits the use of this method on datasets that contain object masks. CenterNet~\cite{zhou2019objects} showed performance improvement by detecting the center as well as the size of the object. This method also shares many similarities with DenseBox~\cite{HuangYDY15}, GuidedAnchoring~\cite{Wang_2019_CVPR}, FAFS~\cite{Zhu_2019_CVPR}, FCOS~\cite{Tian_2019_ICCV}, and FoveaBox~\cite{kong2019foveabox} where they output two types of maps: (i) a confidence (heatmap) per object class predicting the center of objects, and (ii) a shape map predicting the bounding boxes. Instead of directly predicting a heatmap of centers, we propose to utilize the field vectors to point to object centers. Objects in aerial images experience huge variations in orientation compared to general object datasets (MS-COCO~\cite{lin2014microsoft}). We solve this problem by having many output feature cells point towards the center, decreasing the reliance on specific features such as only the center of an object. To take into account the consensus of all vectors, we develop a voting mechanism to locate the center of an object. This mechanism provides us with a two fold advantage: we not only suppress overlapping predictions, but also accumulate the information thereby providing a more refined output.

Instead of directly predicting a heatmap of centers, we use vectors from different output cells to point to object centers. A similar idea is used for instance segmentation~\cite{Kendall_2018_CVPR,8500621,Neven_2019_CVPR,Cheng_2020_CVPR} where the vectors predicted from different output cells is used to cluster related pixels. In contrast, rather than using these vectors to group feature cells, we make explicit use of where these vectors are pointing to localize the center of an object. We thus develop a voting mechanism to refine our predictions by allowing different vectors to agree to a final prediction, allowing us to deal with occlusion as well as scale variations.

\section{Method}

The goal of our method is to detect objects (such as cars, buses, and pedestrians) in aerial images. We propose an anchor-free method that overcomes the limitations of current anchor-based and anchor-free methods. Existing anchor-free detectors output a sparse heatmap wherein each cell is in charge of only locally detecting either a corner or a center of a bounding box. This is typically challenging for a network due to the fact that it needs to output a low confidence even for features that contain information about the object but do not lie at the specific corner of interest. Thus, we extend fields from 2D pose estimation to object detection and introduce \textbf{Butterfly Fields} which are named based on the shape they form when drawn over an image (Figure \ref{fig:pull}).

\begin{figure*}[t!]
\begin{center}
 \includegraphics[width=\linewidth]{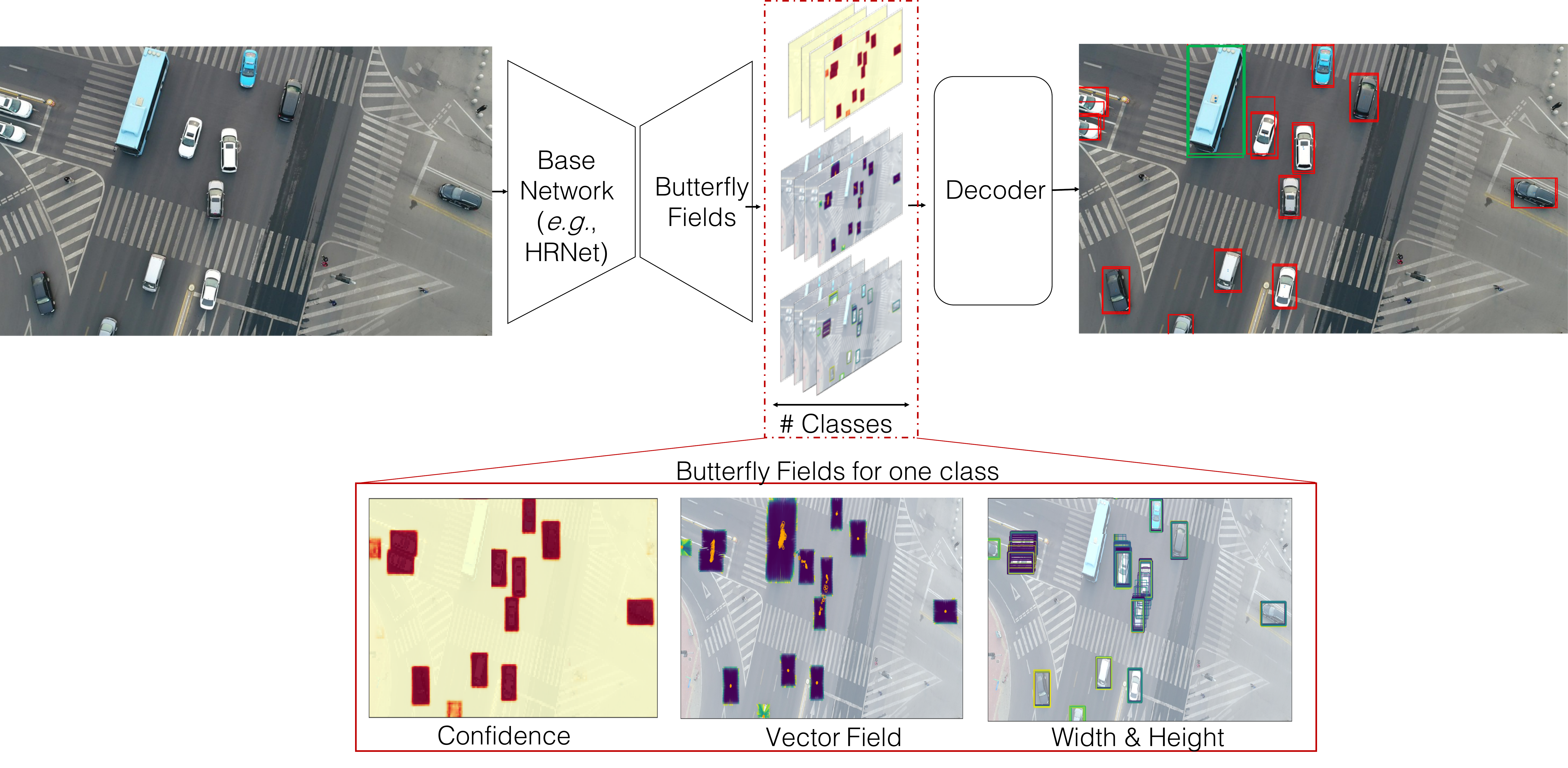}
\end{center}
\caption{Model architecture showing the different components of our butterfly fields. The features extracted by the base-network are used by the head responsible to output the butterfly fields. The top component, represented by a heatmap over the input, indicates the confidence scalar component. The middle component is made up of vectors where each activated feature cell is pointing towards the center (orange circle). The bottom component shows the width and height predicted by each activated output cell. The resulting fields are then decoded to output the final bounding box predictions (right image)}
\label{fig:systemfigure}

\end{figure*}

Figure~\ref{fig:systemfigure} illustrates the overall method. A base network, such as HRNet~\cite{Sun_2019_CVPR}, is used to extract features from an aerial image. These features are then processed by the head responsible in outputting the composite fields, made up of a confidence, width and height of the bounding box, as well as vectors pointing towards the center of every object. In the following sections, we highlight the main parts of our method and the advantage of using fields for object detection compared to previous methods.

\subsection{Butterfly Fields}

General anchor-free object detectors usually assign a single feature cell to be responsible to detect the center of the object by predicting a confidence as well as the offset of the center relative to this cell~\cite{law2019cornernet,wang2017point}. Some methods also detect other attributes such as the width and height~\cite{zhou2019objects}. This forces the neural network to output a sparse feature map with very few and specific activated location while reducing all cells around them. This task is extremely challenging in high-resolution images, where many cells may contain similar information. To overcome this drawback, we introduce \textbf{Butterfly Fields}. The goal of these fields is to localize the center of the object, width, and height from every feature cell that contains information about the object of interest. The difference between other anchor-free methods is explained in Figure~\ref{fig:comparison}. Predicting the center, height, and width, instead of corners does not require association between the corners, which would make inference slower. For instance, if N top-left corners and N bottom-right corners were detected, this requires N$^2$ computations to cluster the corners and form the bounding boxes.

Butterfly fields are composite fields made up of a scalar component for confidence, a vector component that points toward the closest center of an object, and two scalar components indicating the width and height of the predicted object. This can be represented over the feature map \textit{f} as:
\begin{equation}
f^{c}_{ij} = \{p^{c}_{ij}, x^{c}_{ij}, y^{c}_{ij}, w^{c}_{ij}, h^{c}_{ij}\}
\label{eq:onefield}
\end{equation}
where $i$ and $j$ index the output feature cell, and \textit{c} represents the object class. $p$ is the confidence of the cell in its prediction, $x$ and $y$ form the vector from the center of the cell to the center of the object. $w$ and $h$ are the the width and height respectively of the referred object. In the case of overlapping objects of same class, the vector components will point to the closest center since the datasets used do not have any information about relative object positions.

As can be seen from Equation~\ref{eq:onefield}, a composite field is predicted for every object category. This allows the network to output different fields that are specialized for specific objects with specific aspect ratios. For instance, a model can learn that a pedestrian has a specific width to height ratio and thus the output feature map specific for this class would capture this characteristic. On the contrary, previous keypoint-based object detectors predict a common width, height, and even offset for all categories. Our modification allows us to obtain specialized predictors for different classes.

\subsection{Loss functions}
In order to train the model, a ground-truth feature map is built based on the bounding box annotations provided by the dataset. At every feature cell inside a bounding box, the butterfly field components are set such that the confidence $p$ is 1. $w$ and $h$ are set to be the logarithm of the down-scaled width and height respectively in Eq~\ref{eq:onefield}. The vector component ($x$ and $y$) are set to point to the center of the object from the center of the corresponding feature cell. All other feature cells that do not contain any object have their confidence component $p$ set to zero. We also compare the performance of our model when different number of cells around the center are chosen to predict the bounding box. For instance, we test on our model when using a 4x4 window around the center
Figure~\ref{fig:sub-cbbox16} and when using all feature cells inside the bounding box Figure~\ref{fig:sub-cbboxfullfields}.

Two main losses are used in order to train our network. The scalar components that represent the confidence are trained with independent binary cross-entropy, while the components of the field related to the width and height are trained using an L1 Loss. To train the vector components of the butterfly field, we make use of the Laplace Loss~\cite{kendall2017uncertainties,Bertoni_2019_ICCV}:
\begin{equation}
    L = \frac{|x-\mu|}{b} + \log{2b}
\end{equation}
The model learns $b$ without any supervision and is used to attenuate the gradients based on the model's confidence in its predictions. It is important to note that we do not deal with the imbalance of negative and positive samples compared to other anchor-free object detectors that make use of a modified focal loss~\cite{abs_1708_02002}. The focal loss was tested and did not show any improvement compared to using the normal binary cross-entropy.

\subsection{Decoding: Consensus of every point in a field}\label{sec:decoding}

During inference, given the fields over the output feature, we extract the set of bounding boxes with their confidences and appropriate classes. In order to deal with the huge variation of objects, we develop a decoding procedure that scales according to the height and width of every object.

For every class-specific feature output, we build a high resolution confidence map using both the confidence and vector components of the butterfly fields. This produces a heatmap over the input image where the intensity of a region corresponds to its probability of being the center of an object. 
We propose the following adaptive function to build our heatmap given the output feature map:
 \begin{equation}
 \small
    f(x,y) =
    \sum_{ij}\frac{p_{ij}}{\chi_{ij}}
    \exp \left(
        -\frac{1}{2}
            \left(\frac{x - x_{ij}}{\sigma^{x}_{ij}}\right)^{2}
        -\frac{1}{2}
            \left(\frac{y - y_{ij}}{\sigma^{y}_{ij}}\right)^{2}
    \right)
\label{eq:gaussian}
\end{equation}
where $x$ and $y$ index a pixel in the high-resolution confidence map, and $i$ and $j$ vary over the output feature cells with confidences greater than $0.1$. Each output cell predicts a center of an object ($x_{ij}$, $y_{ij}$) and confidence $p_{ij}$. Instead of adding the confidences only at the predicted center, Equation~\ref{eq:gaussian} uses an unnormalized 2D Gaussian to spread this confidence in a neighborhood of size $\sigma_{ij}$. A variable number of output cells can predict the same object center and thus we need to deal with the variable normalization of the accumulated confidence. Since summing all the
$p_{ij}$ in Equation~\ref{eq:gaussian} will result in confidences greater than 1, we use $\chi_{ij}$ to normalize the confidence based on the number of expected vectors voting for the same center. This value scales with the number of cells localizing an object center (related to the number of output cells inside the bounding box). In the case of using 16 cells around the center (4x4 window), $\chi_{ij}=16 \ \forall i,j$. When using all the feature cells inside the bounding box to predict the center, $\chi_{ij}$ is the area of the corresponding box. We define $\sigma^{x}_{ij}$ and $\sigma^{y}_{ij}$ as functions of the width and height respectively:

\noindent\begin{minipage}{.5\linewidth}
\begin{equation}
    \sigma^{x}_{ij} = \max\left(2, \frac{w_{ij}}{\rho}\right)
\label{eq:sigma_x}
\end{equation}
\end{minipage}%
\noindent\begin{minipage}{.5\linewidth}
\begin{equation}
    \sigma^{y}_{ij} = \max\left(2, \frac{h_{ij}}{\rho}\right)
\label{eq:sigma_y}
\end{equation}
\end{minipage}\\
These equations can be interpreted as the maximum number of allowed error pixels for predicted bounding box center. This depends on the size of the object since the error in prediction for small objects is more costly than the prediction for big objects. $\rho$ is a hyperparameter that defines the amount of error allowed relative to the size of the object. It can be a single number or a distinct number for each class. We set the minimum value for $\sigma$ to be 2 pixels (shown in Equation~\ref{eq:sigma_x} and ~\ref{eq:sigma_y}).

Based on this voting mechanism, the resulting high-resolution confidence map contains peaks highlighting the probability of the existence of object centers. In addition, another consensus occurs when calculating both the width and height. They are calculated using a weighted average of all predictions following this equation:
\\
\noindent\begin{minipage}{.5\linewidth}
\begin{equation}
    w(x,y) = \frac{\sum_{ij}{p_{ij}w_{ij}}}{\sum_{ij}{p_{ij}}}
\end{equation}
\end{minipage}%
\noindent\begin{minipage}{.5\linewidth}
\begin{equation}
    h(x,y) = \frac{\sum_{ij}{p_{ij}h_{ij}}}{\sum_{ij}{p_{ij}}}
\end{equation}
\end{minipage}\\
where $i$ and $j$ are the indices of all feature cells that have their vector fields pointing towards a specific feature located at $x$ and $y$.

\begin{figure*}[t]

\begin{subfigure}{0.3\linewidth}
  \centering
   \includegraphics[width=\linewidth]{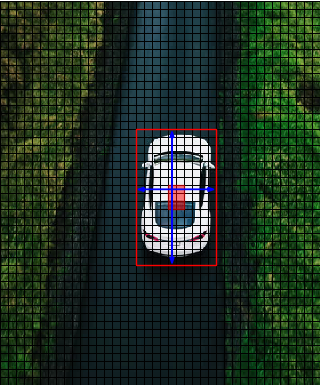}
  \caption{ }
  \label{fig:sub-previous}
 \end{subfigure}%
 \hspace{4mm}
 \begin{subfigure}{0.3\linewidth}
  \centering
  \includegraphics[width=\linewidth]{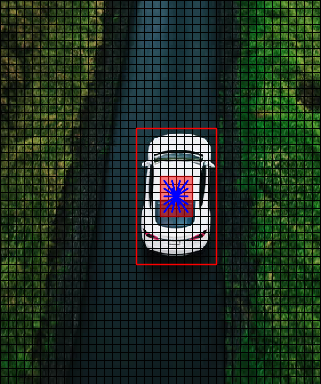}
  \caption{ }
   \label{fig:sub-cbbox16}
 \end{subfigure}%
 \hspace{4mm}
 \begin{subfigure}{0.3\linewidth}
  \centering
  \includegraphics[width=\linewidth]{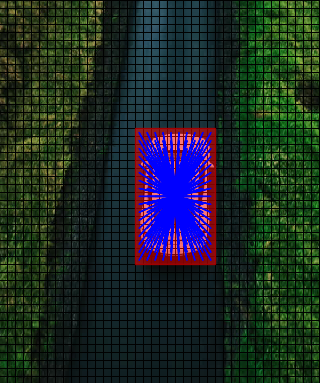}
  \caption{ }
   \label{fig:sub-cbboxfullfields}
 \end{subfigure}
\caption{A visualization of the difference between anchor-free methods (a)~\cite{HuangYDY15,Wang_2019_CVPR,Zhu_2019_CVPR,Tian_2019_ICCV,kong2019foveabox} and our buttefly detector (b-c). (b) shows our method using a 4x4 window around to point towards the center. (c) shows our method using all grid cells inside the bounding box. Instead of having a map indicating the center, we make use of fields over the image where each vector is pointing towards the center to localize it. The voting mechanism aggregates information from multiple cells to obtain a refined prediction}
\label{fig:comparison}

\end{figure*}

\begin{figure*}[t]

\begin{subfigure}{0.5\linewidth}
  \centering
   \includegraphics[width=\linewidth]{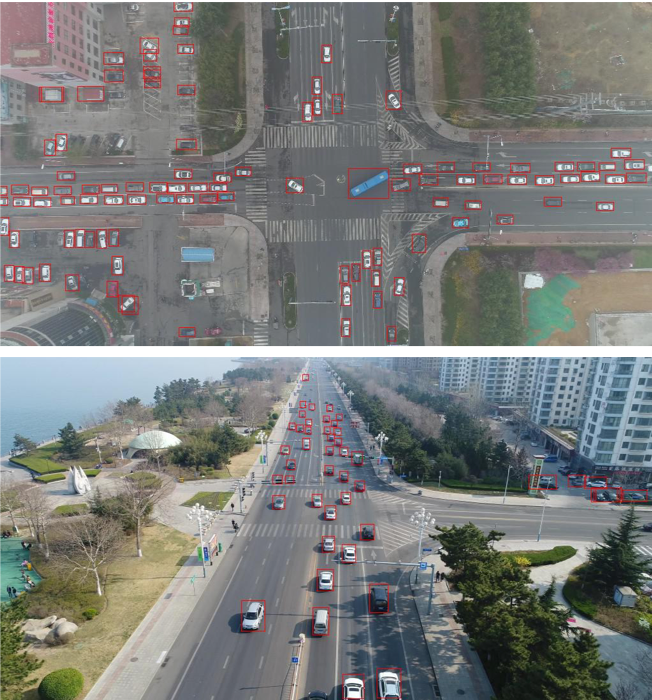}
   \caption{UAVDT}
  \label{fig:sub-first}
 \end{subfigure}%
 \hspace{5mm}
 \begin{subfigure}{0.47\linewidth}
  \centering
  \includegraphics[width=\linewidth]{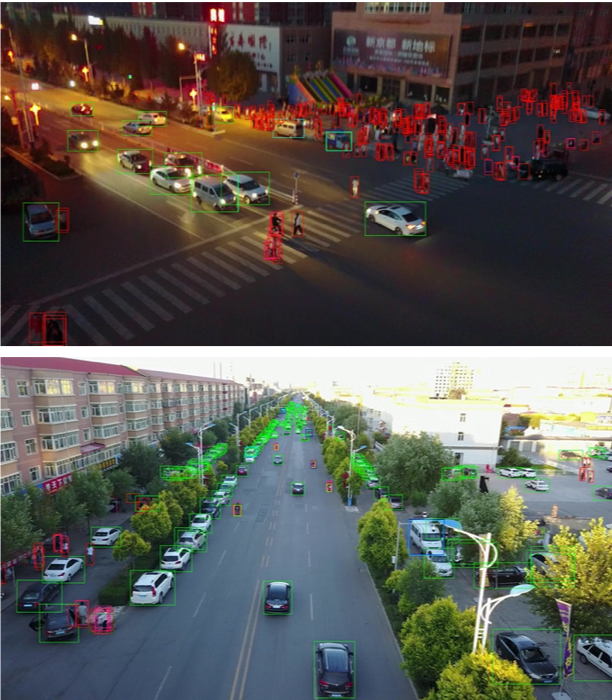}
  \caption{VisDrone2019}
   \label{fig:sub-second}
 \end{subfigure}
\caption{Qualitative results of the Butterfly Detector on UAVDT and VisDrone dataset. Our method was able to perfectly detect partly occluded objects (\textit{e.g.}, images from UAVDT). This is because our Butterfly Detector leverages several locations in a field pointing to the center through a voting mechanism}
\label{fig:qualitative}

\end{figure*}

Figure~\ref{fig:comparison} illustrates the two key differences of our method with respect to previous works.
Firstly, the confidence map of previous methods is usually dense only around the center of the object (Figure~\ref{fig:sub-previous}). Such a sparse feature map results in difficulty in training. To tackle this, our method utilizes a greater number of feature cells within each bounding box.  Each feature cell falling inside the bounding box predicts a vector pointing toward the object center (Figure~\ref{fig:sub-cbbox16}-\ref{fig:sub-cbboxfullfields}). Moreover, the feature cell confidence $p$ indicates the confidence of the pointed vector rather than the probability of this cell being the center of an object. A map is also built where each feature cell predicts the height and width at the location pointed by the vector. Secondly, previous methods rely on Non-Maximum Suppression to filter their predictions. Instead, we propose our voting mechanism that allows the network to overcome many challenges such as occlusion. Consider the case in which a confidence map detecting the center is used, such as in CenterNet~\cite{zhou2019objects}; if more than half of the object of interest is occluded, the method would find difficulties detecting the center. Since we leverage cues from different feature cells to detect the center, we are able to overcome such shortcomings. For instance, by detecting only the trunk of a car, our method is able to correctly output a tight bounding box as shown in the qualitative results (Figure~\ref{fig:qualitative}).

\subsection{Decoding: Sub-pixel Localization}

\begin{figure*}[t]
\begin{subfigure}{0.5\linewidth}
  \centering
   \includegraphics[width=0.55\linewidth]{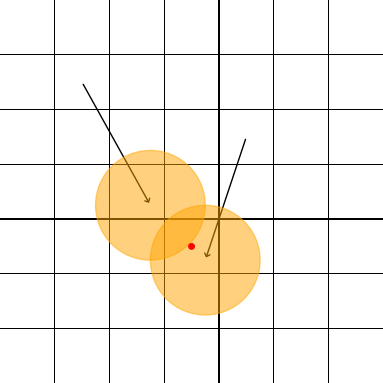}
  \caption{ }
  \label{fig:sub-normalpixel}
 \end{subfigure}%
 \hspace{4mm}
 \begin{subfigure}{0.5\linewidth}
  \centering
  \includegraphics[width=0.55\linewidth]{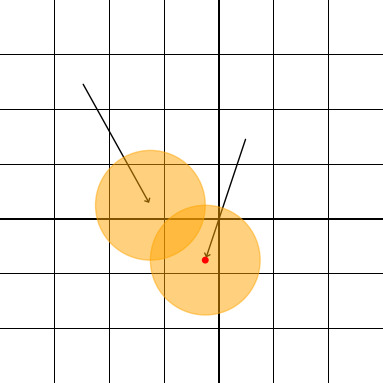}
  \caption{ }
   \label{fig:sub-subpixel}
 \end{subfigure}%
\caption{A visualization comparing using sub-pixel decoding (b) to using directly the high resolution map (a). The yellow circles represent the Gaussian used to fill the high-confidence resolution map (Eq.~\ref{eq:gaussian}). The red circle is the resulting object center since it should be located in the cell with a high confidence}
\label{fig:comparison_subvspixel}
\end{figure*}

Upon building a high-resolution confidence map, cells with high confidence represent the center of objects. A loss of spatial precision occurs when using the center of cells as the centers. As can be seen in Figure~\ref{fig:sub-normalpixel}, using the center of the cell with the most Gaussian overlap might lead to an average between two predictions. This will lead to a loss of precision and will affect the detection of small objects. To remedy this issue, we make use of the floating-precision of our vector components. Instead of taking the center of the highest confident cell as the center of an object (red dot in Figure
~\ref{fig:sub-normalpixel}), we take the exact location the vector is pointing to inside this cell (red dot in Figure
~\ref{fig:sub-subpixel}). This method leads to a boost in performance, especially for the VisDrone2019 since it contains many small objects, such as people and bicycles (see Section \ref{sec:exp}).

\section{Experiments}\label{sec:exp}
To evaluate our method, we study the performance of the butterfly detector on two publicly available UAV datasets: UAVDT~\cite{du2018unmanned} and VisDrone2019~\cite{zhuvisdrone2018}. In order to showcase the importance of the different parts of our method, we also perform a detailed ablation study.


Our experiments show the advantage of using composite fields for object detection. Combined with a high resolution confidence accumulation procedure and a sub-pixel localization, we demonstrate state-of-the-art-performance for object detection in aerial images.
\subsection{Datasets}

\paragraph{UAVDT} is a dataset collected using a UAV platform at different locations, altitudes, weather conditions, as well as viewing angle. This makes it a challenging dataset for both tracking and detection. The dataset also includes annotations for the previously listed variations in addition to the amount of occlusion. It contains three different classes: car, truck, bus. The detection aspect of this dataset is made up of around 41k frames from 50 distinct videos, 20 of which are used for testing.
\paragraph{VisDrone 2019} is a new dataset that is also collected using a UAV platform. Compared to UAVDT, the detection aspect of this dataset is relatively small but contains much more categories. The 8.5k images include 10 classes: pedestrian, people, bicycle, car, van, truck, tricycle, awning-tricycle, bus, and motor. The small number of images and big number of categories introduces another challenge for our method. Moreover, the difference between pedestrian and people category is that pedestrians are people who are walking or standing still.

\subsection{Implementation and Training Details}
Our model uses an ImageNet pre-trained HRNet as a base-network since it recently showed good performance for keypoint estimation, a task that shares some similarities to our method.  We augment the base-network with a head composed of 1x1 convolutional layers responsible to output the different components of our butterfly fields.

During training, we apply different standard data augmentation techniques. After normalizing the images, we randomly flip the input horizontally. The resulting image is then rescaled by a scale between 0.4 and 2.0. We subsequently extract a random 512x512 crop that also undergoes a random $90^{\circ}$ rotation.

As mentioned previously, the confidence map is trained using a binary cross entropy loss. The vector components of our fields are trained using a Laplace loss while all other scalar components, such as width and height, are trained using an L1 loss. In order to find the best prefactor dictating the contribution of each loss on the gradient, we make use of the work by Kendall et al.~\cite{Kendall_2018_CVPR} that weighs the different losses based on their homoscedastic uncertainty, thereby reducing the number of hyperparameters. The model is trained using the SGD optimizer, a momentum of 0.95 and a learning rate of $2x10^{-3}$. The batch size is set to 5, and the training runs for 70 epochs for UAVDT and 150 epochs for VisDrone2019. In the cases where not all the cells inside the bounding box are used to localize the center, a region defined as an "ignore" region, with a size set to 20\% that of the object in question, is created around the used cells.

\subsection{Testing Details}
During testing, the image is rescaled to have the width and the height as multiples of 32, a requirement based on the architecture of HRNet. The confidence threshold for selecting centers is 0.05, similar to previous methods. We set the $\rho$ to 10 for the UAVDT Dataset and 5 for Visdrone. The latter requires a smaller $\rho$ because it contains small objects. Once the butterfly fields are extracted, we use the voting mechanism introduced in Section~\ref{sec:decoding} to aggregate and obtain the predicted bounding boxes. The resulting bounding boxes are then refined using soft non-maximum suppression~\cite{Bodla_2017_ICCV}.

Both datasets are evaluated using average precision (AP). UAVDT benchmark requires the AP to be calculated at a specific IOU threshold of 0.7. On the contrary, VisDrone2019 requires an evaluation protocol similar to MS-COCO~\cite{lin2014microsoft} where the average precision as well as average recall are calculated at different IOU thresholds. These metrics are shown in Table~\ref{table:sota_visdrone} and are explained in detail in~\cite{lin2014microsoft}.
\section{Results}

\begin{table}[t]
\caption{Comparison of AP (\%), False Positives (FP), and Recall (\%) with state-of-the-art methods on UAVDT datasets. We outperform previous methods on medium to high altitude images as well as bird-view images. These are significant challenges that general object detectors face on aerial images. BD$_{NV}$ is our method with no voting mechanism and using all predicted bounding boxes. The subscript refers to the number of feature cells used to localize the center. BD$_{full}$ makes use of all feature cell inside the ground truth bounding box. *: trained and tested by us. $fps$: frames per second}
\centering
\ra{1.3}
\resizebox{\textwidth}{!}{\begin{tabular}{@{}llrrcrcrrrcrrrcrrr@{}}\toprule
&& \multicolumn{1}{c}{}&\multicolumn{1}{c}{}& \phantom{a}&\multicolumn{1}{c}{} & \phantom{a}& \multicolumn{3}{c}{Flying Altitude} & \phantom{a}& \multicolumn{3}{c}{Camera View} &
\phantom{a} & \multicolumn{2}{c}{Time} &\\

\cmidrule{8-10} \cmidrule{12-14} \cmidrule{16-17}
& Backbone&FP&Recall&&$avg.$ &&$low $ & $medium$ & $high$ && $front$ & $side$ & $bird$ && $day$ & $night$ & $fps$\\ \midrule
 YOLOv3\cite{redmon2016you}  & DarkNet-53 +FF&--&--&&20.6 && -- & -- &  -- && -- & -- & -- && -- & --&--\\
 SSD~\cite{liu2016ssd} & VGG-16&2,111k&49.28&& 33.6 && -- & -- & -- && -- & -- & -- && -- & --& 42\\
 CADNet~\cite{duan2019detecting} &VGG-16 +FF&--&--&& 43.6 && -- & -- & -- && -- & -- & -- && -- & --&--\\
Faster R-CNN~\cite{ren2015faster} &ResNet-101 &--&--&& 45.64 && 68.14 & 49.71 & 18.70 && 53.34 & 68.02 & 27.05 && 45.63 & 52.14& 2.8\\
GANet~\cite{cai2019guided} &ResNet-50 +FF&--&--&& 47.2 && -- & -- & -- && -- & -- & -- && -- & --&--\\
 NDFT F-RCNN~\cite{wu2019delving} &ResNet-101 &--&--&& 47.91 && \textbf{74.84} & 56.24 & 20.55 && \textbf{64.88} & \textbf{67.50} & 28.79 && 45.91 & 64.16&--\\
 CenterNet$^{*}$[47]& Hourglass-104 &1,204k&66.0&& 53.95 && 70.21 & 60.22 & 27.65 && 57.13 & 66.29 & 38.85 && 62.63 & 67.24&6\\
\hline

BD$_{16}$ & ResNet-50 &453k&62.6&& 52.74&& 68.32 & 61.87 & 21.44 && 57.44 & 65.15 & 34.55 && 62.58 & 69.27&12.2\\

BD$_{NV}$ &HRNet-W32& 2,881$k$ & \textbf{71.28}&& 45.02 &&58.21 & 51.79 & 21.52 && 48.83 & 50.62 & 32 && 50.29 & 64.61&--\\

  BD$_{1}$ & HRNet-W32& 412$k$  & 67.46 &&51.44&& 63.19 & 56.3 & 29.57 && 49.91 & 62.13 & 42.14 && 58.07 & 62.94&18.4\\

BD$_{full}$ &HRNet-W32&  224$k$ & 63.74 &&54.28 && 67.6 & 59.77 & 31.26 && 55.16 & 63.88 & 42.79 && 60.84 & 70.88&15.4\\

BD$_{16}$ &HRNet-W32& 313$k$ &69.46&& \textbf{58.1} && 67.84& \textbf{61.93} & \textbf{40.6} && 56.28 & 66.81 & \textbf{51.57} && \textbf{62.97} & \textbf{69.59}& \textbf{18.4}\\

  \bottomrule
\end{tabular}}
\label{table:sot_uavdt}

\end{table}

\subsection{Comparison to State-of-the-art}

Our method is able to significantly outperform previous state-of-the-art methods on UAVDT dataset. As seen in Table~\ref{table:sot_uavdt} and comparing it to published state-of-the art method(NDFT F-RCNN), our butterfly detector reaches an average precision of 58.1\%, exceeding it by $\sim 10\%$. We also validate our motivation for the use of fields in aerial images. This is shown by the drastic improvement of $\sim 14\%$ in average precision on medium to high altitude images as well as an improvement of $\sim 17\%$ on images taken from bird view. We also train a state-of-the-art anchor-free object detector, CenterNet, on UAVDT. It uses an Hourglass architecture with input size 512x512. Testing is done on the full resolution. Our method outperforms CenterNet by $\sim 4\%$ (Table~\ref{table:sot_uavdt}) while obtaining similar results using ResNet-50. Compared to CenterNet, we especially achieve better results on medium and high altitude images (+1.7\% and +13\% respectively) as well as on bird-view images ($\sim +13\%$)

\begin{table*}[t]
\caption{Comparison of AP (average precision), AR (average recall), True Positives (TP), and False Positives (FP) with state-of-the-art methods on VisDrone dataset. As shown, we outperform previous methods on all metrics. $det.$: specifies the maximum detections chosen. **: trained and tested by us. $\star$: use of sub-pixel localization}
\centering
\ra{1.3}
\resizebox{\textwidth}{!}{\begin{tabular}{@{}lrrrcrrrrcrcr@{}}\toprule
& \multicolumn{3}{c}{AP(\%, det.=500)} &
\phantom{a} & \multicolumn{4}{c}{AR(\%, IoU$_{0.5:0.95}$)} &\phantom{a}& \multicolumn{1}{c}{TP} &\phantom{a}& \multicolumn{1}{c}{FP} \\

\cmidrule{2-4} \cmidrule{6-9} \cmidrule{11-11} \cmidrule{13-13}
& IoU$_{0.5:0.95}$ & IoU$_{0.5}$ & IoU$_{0.75}$ && $det.=1$ &$det.=10$ & $det.=100$ & $det.=500$&& IoU$_{0.75}$&& IoU$_{0.75}$\\
\midrule
  FRCNN\cite{ren2015faster} +FPN\cite{lin2017feature} & 21.4 & 40.7 & 19.9 && -- & -- & -- & --&& --&& --\\
  CenterNet~\cite{law2018cornernet}* & 22.96 & 42.05 & 21.81 && 0.68 & 6.14 & 32.56 & 32.56 && 15.9k&& 36k\\
  ClusDet~\cite{abs_1904_08008} & 28.4 & 53.2 & 26.4 && -- & -- & -- & -- && --&& --\\
   \hline

   BD & 28.93 & 54.92 & 26.15 && 0.98 & 7.06 & 36.1 & 40.93&& 17.8k&& 46.7k\\

   BD$^{\star}$ & \textbf{30.15} & \textbf{54.67} & \textbf{28.66} && \textbf{0.98} & \textbf{7.11} & \textbf{37.44} & \textbf{41.41}&& 19.3k && 31.7k\\

  \hline
\end{tabular}}

\label{table:sota_visdrone}

\end{table*}

We also obtain state-of-the-art results on the VisDrone2019 dataset (Table~\ref{table:sota_visdrone}). We report our results on the validation set. Our detector is able to attain better performance than all previous methods. ClusDet~\cite{abs_1904_08008}, a two-stage detector, achieves good results since it runs inference on multiple small patches of the image as well as the full image. Although this allows it to be more fine-grained, we outperfom it by $\sim 2.2\%$. Comparing it to CenterNet~\cite{zhou2019objects}, an anchor-free object detector that suffer from the limitations discussed previously, our method outperforms it by $\sim 7\%$ when averaging over the different IoUs and by $\sim 12\%$ at IoU=0.5. The improvement of the AP at IoU=0.75 indicates that our method is able to correctly predict the scale of the objects.

\subsection{Inference Speed}

We also focus on the inference time of previous methods and compare them to our butterfly detector (Table~\ref{table:sot_uavdt}). Most methods did not provide their inference time for comparison. As observed, Faster R-CNN has a low fps due to it being a two-stage detector that relies on a region-proposal network(RPN). Since the architecture and base network of NDFT F-RCNN is based on Faster R-CNN, it is expected that it also does not run in real-time reaching around $~3\ fps$. This is also the case for CADNet that also makes use of a more complicated RPN. Our method performs better than all previous state-of-the-art while running at a speed of 18.4 fps, a real-time performance. All reported inference speeds were obtained using a GTX 1080 Ti.

\section{Ablation Study}

\paragraph{Effect of Fields.} In order to showcase the benefits of fields, we compare in Table~\ref{table:sot_uavdt} our  method (BD$_{16}$) to the same method but using only one feature cell responsible for the center (BD$_{1}$). As can be observed, a gain of $\sim$ 7\% is achieved when different locations in a field are used to vote for the localization of the center instead of only using the center cell. We observe that we achieve a boost of $2\%$ in recall as well. This is due to the fact that different bounding boxes are being proposed based on the extracted features at different output cells.

\paragraph{Effect of Voting Mechanism.}
To study the advantages of voting, we measure the performance of the model when all the bounding boxes predicted by the activated output feature cells are used. Our method without the voting mechanism achieves an AP of 45.02\% (Table~\ref{table:sot_uavdt}) which is almost on-par with Faster R-CNN, a slower two stage detector. As expected, we observe a large difference between true positives and false positives leading to a high recall but low average precision. When the voting mechanism described in section~\ref{sec:decoding} is applied, a significant decrease in the false positives is achieved while preserving the recall. As a result, a significant boost in the AP is observed of around 13\% increasing from 45.02\% to 58.1\%. This improvement is also shown over all the different attributes of the UAVDT dataset.

\paragraph{Effect of Backbone.} In order to better verify the positive effect of our method, we replace the HRNet-W32 with a commonly-used backbone, ResNet-50. Our method with a ResNet-50 architecture is able to outperform previous methods while achieving similar results to CenterNet. These previous methods make use of larger architectures as well as fusion between different feature scales (Table~\ref{table:sot_uavdt}). HRNet-W32 is used in our method since it allows learning from different feature scales similar to CenterNet's backbone, Hourglass-104.

\paragraph{Number of Localizing Cells in a Field.} We also study the effect of the number of points in a field localizing the center of an object on the performance. We evaluate the performance of our method while using one feature cell which is the center (BD$_1$), 16 feature cells around the center (BD$_{16}$), and all feature cells lying inside the bounding box (BD$_{full}$). Table~\ref{table:sot_uavdt} shows the numbers of true and false positives as well as the average precision per attribute for these three configurations. We notice that the best settings is using 16 cells in a 4x4 window to localize the center and predict the width and the height of the bounding box. BD$_{full}$ works by utilizing all features inside the bounding box. Since ground truth boxes are axis-aligned, a bounding box of a car moving diagonally relative to the image will contain areas of the road. In such cases, using the full boxes requires classifying some roads with high confidence leading to a difficult training process. Thus, this can explain the slightly lower performance compared to BD$_{16}$. Nonetheless, our detector with the full bounding box still outperforms previous methods (Table
~\ref{table:sot_uavdt}). As for BD$_{1}$, it makes use of only the center feature to localize the object which makes it challenging to train and decode as explained previously.

\paragraph{Effect of Sub-pixel Localization.} To better observe the effect of the sub-pixel localization, we perform this ablation study on VisDrone2019. As observed in Table~\ref{table:sota_visdrone}, using the sub-pixel technique improves the performance of the model by $\sim 1.2\%$. This method also helped in drastically reducing the false positives while also increasing the true positives.

\section{Conclusions}

In this paper, we propose a new object detector, Butterfly Detector, that utilizes composite fields over the spatial feature map to localize the center of objects and predict their sizes. Using the extracted fields for an object center, we apply a voting mechanism to further improve our detector especially in cases of occlusion or viewing angle variations. We evaluated our approach on UAVDT and VisDrone and show that it outperforms the previous state-of-the-art methods. We further perform an ablation study to demonstrate the benefits of butterfly fields, coupled with a voting process and sub-pixel localization, for object detection in aerial images.


\end{document}